\documentclass{article} % For LaTeX2e
\usepackage{iclr2027_conference,times}

\usepackage{amsmath,amsfonts,bm}

\def\eqref#1{equation~\ref{#1}}
\def\1{\bm{1}}

\DeclareMathAlphabet{\mathsfit}{\encodingdefault}{\sfdefault}{m}{sl}
\SetMathAlphabet{\mathsfit}{bold}{\encodingdefault}{\sfdefault}{bx}{n}

\usepackage{hyperref}
\usepackage{url}
\usepackage{graphicx}
\usepackage{comment}

\usepackage{amsmath} % For \boxed
\usepackage{booktabs}
\usepackage{multirow}
\usepackage{makecell}
\usepackage{graphicx}

\usepackage{amssymb}
\usepackage{placeins}
\usepackage{dblfloatfix}
\usepackage{tabularx}
\usepackage{array}
\usepackage{newfloat}
\usepackage{listings}
\usepackage{caption}
\usepackage{booktabs}
\usepackage{multirow}
\usepackage{makecell}
\usepackage{adjustbox}
\usepackage{graphicx}
\usepackage{hyperref}
\usepackage{url}
\usepackage{graphicx}
\usepackage{comment}
\usepackage{wrapfig}
\usepackage{amssymb}
\usepackage{amsmath} % For \boxed
\usepackage[table]{xcolor}
\usepackage{tikz}

\definecolor{blueLight}{HTML}{EBF5FF}
\definecolor{resourceBlue}{HTML}{00549F}
\hypersetup{colorlinks=true,linkcolor=black,citecolor=blue,urlcolor=resourceBlue}

\makeatletter
\def\section{\@startsection {section}{1}{\z@}{-2.0ex plus
    -0.5ex minus -.2ex}{1.5ex plus 0.3ex
minus0.2ex}{\large\sc\raggedright\color{resourceBlue}}}
\makeatother

\renewenvironment{abstract}{%
\par\vspace{0.6em}%
\noindent\begin{tikzpicture}%
\node[fill=blueLight,rounded corners=10pt,inner sep=14pt] \bgroup%
\begin{minipage}{\dimexpr\textwidth-28pt\relax}%
\textbf{Abstract.}\ }{%
\end{minipage}%
\egroup;%
\end{tikzpicture}%
\par\vspace{1ex}%
}

\usepackage{newfloat}
\usepackage{listings}
\usepackage{caption}

\title{ActionSplice: In-Flight Action Editing\\
for Interactive World Models}

\author{
\begin{tabular}{c}
Pardis Taghavi \quad Tingyu Guo \quad Jonas Lossner \quad
Gaurav Pandey \quad Reza Langari \\
\normalfont Texas A\&M University \\[0.5em]
\normalfont
\href{https://github.com/PardisTaghavi/ActionSplice}{\textcolor{resourceBlue}{\textbf{Code}}}
\quad
\href{https://huggingface.co/PardisTaghavi/ActionSplice}{\textcolor{resourceBlue}{\textbf{Checkpoints}}}
\quad
\href{https://pardistaghavi.github.io/actionsplice-website/}{\textcolor{resourceBlue}{\textbf{Project Page}}}
\end{tabular}
}

\iclrfinalcopy % Uncomment for camera-ready version, but NOT for submission.
\begin{document}

\pagestyle{plain}
\maketitle

%Chunk-autoregressive video world models typically condition each generated chunk on a single action, forcing control updates to wait for the next chunk and limiting action frequency to the chunk-generation rate.

\vspace{-1.7em}
\begin{abstract}
Chunk-autoregressive video world models typically generate each chunk under one action. When an action changes during sampling, waiting until the next chunk delays the response, while directly switching the conditioning leaves the intermediate solver state shaped by the previous action. Restarting sampling under the revised action avoids this mismatch but repeats completed computation. We introduce \emph{ActionSplice}, which edits the interrupted state through Counterfactual State Transport (CST). A lightweight corrector moves the interrupted solver state toward the matched counterfactual solver state induced by the revised action at the same solver step. The world model and sampler remain frozen, and sampling resumes without replaying completed evaluations. $\mathrm{CST}_{R}$ retargets the entire active chunk, while $\mathrm{CST}_{T}$ preserves the temporal prefix at the intervention step and corrects only the suffix. On minWM--Wan Action2V and HY-WM1.5, we measure fidelity against matched rollback references. Relative to condition swapping, $\mathrm{CST}_{R}$ reduces LPIPS by 61.5\% and 75.9\%, and $\mathrm{CST}_{T}$ reduces suffix LPIPS by 56.1\% and 77.5\%, respectively.

\end{abstract}

%\section{Introduction}
%\label{sec:intro}

\section{Introduction}
\label{sec:introduction}

%The goal is not simply to “learn when to interrupt.” It is to learn how to correct an already partially denoised chunk when a new action arrives, without restarting generation.

Interactive video world models must generate frames quickly and respond promptly to changes in control. Recent diffusion world models support action-conditioned visual simulation and real-time streaming generation \citep{valevski2025diffusion,wang2026matrix,zhao2026minwm}. However, high frame throughput does not guarantee a short delay between an action request and its visible effect. We consider chunk-autoregressive generation in which each chunk is initially sampled under one action through \(K\) solver evaluations. If a revised action arrives after evaluation \(r<K\), the active chunk is only partially generated, but its intermediate state already reflects the previous action. Waiting until the next chunk postpones the response. We study \emph{in-flight action editing}: incorporating the revised action into the active chunk without replaying completed solver evaluations or modifying the committed history of completed chunks.

%When the action changes, the interrupted representation reflects the previous action.  
Direct condition swapping changes conditioning for subsequent solver evaluations but does not modify the representation at the moment of the switch. We refer to this mismatch as \emph{state misalignment}. Full rollback resolves it by restarting the active chunk under the revised action, but repeats the first \(r\) solver evaluations. The representation reached at step \(r\) provides a counterfactual target under matched initial conditions and committed history. We seek to approximate this target from the interrupted representation, allowing sampling to continue without repeating completed evaluations.

We introduce \emph{ActionSplice}, an inference framework built around Counterfactual State Transport (CST). A lightweight corrector trained on
matched rollback pairs predicts a masked residual at the interruption step,
producing a resumable solver state. The frozen world model then performs the
remaining \(K-r\) evaluations. The retargeting variant \(\mathrm{CST}_{R}\) applies the revised action to the entire active chunk. The temporal splicing variant \(\mathrm{CST}_{T}\) supports a scheduled action transition at a temporal boundary \(m\) within the chunk. It preserves prefix coordinates of the interrupted solver state at the intervention step and restricts the correction to the suffix. The solver step \(r\) specifies when sampling is interrupted. The temporal boundary \(m\) specifies where the action changes in the generated sequence. Full rollback constructs training targets and evaluation references. It is not executed during ActionSplice inference.

We evaluate CST on minWM--Wan Action2V~\citep{zhao2026minwm} and
HY-WM1.5~\citep{hunyuanworld2025hy}, with a separate corrector for each
backbone and variant. We examine receipt steps, directed action transitions,
temporal boundaries, and repeated interruptions. We also evaluate camera
trajectories and human judgments of action following and response delay.
Fidelity is measured against matched full rollback for
\(\mathrm{CST}_{R}\) and matched rollback with prefix clamping for
\(\mathrm{CST}_{T}\). Relative to condition swapping,
\(\mathrm{CST}_{R}\) reduces LPIPS by 61.5\% on minWM and 75.9\% on
HY-WM1.5, while \(\mathrm{CST}_{T}\) reduces suffix LPIPS by 56.1\% and
77.5\%, respectively. Compared with waiting, \(\mathrm{CST}_{R}\) achieves
pixel-ready speedups of \(2.69\times\) and \(1.64\times\) on minWM and
HY-WM1.5, respectively. On the HY-WorldPlay benchmark,
\(\mathrm{CST}_{R}\) obtains a PSNR of 25.66 against the original rollout.
\section{Related Work}
\label{sec:related_work}

\par\noindent\textbf{Interactive video world models.}
Diffusion-based world models support action-conditioned visual simulation in interactive game environments~\citep{alonso2024diffusion,valevski2025diffusion}. Recent systems extend these capabilities to streaming generation with camera control. WorldPlay~\citep{sun2025worldplay} uses dual action representations and reconstituted context memory to maintain geometric consistency over long rollouts. Matrix-Game 3.0~\citep{wang2026matrix} combines long-horizon memory with multi-segment autoregressive distillation. HY-World 1.5~\citep{hunyuanworld2025hy} targets real-time interactive generation with geometric consistency. minWM~\citep{zhao2026minwm} and BiWM~\citep{rui2026biwm} provide frameworks for adapting bidirectional video diffusion backbones into autoregressive world models with camera control. Other work addresses the training and efficiency of autoregressive generators. Causal Forcing~\citep{zhu2026causal}, Causal Forcing++~\citep{zhao2026causal}, and Causal-rCM~\citep{zheng2026causal} develop training and distillation methods for efficient causal generation. Self Forcing~\citep{huang2026self} addresses exposure bias by training on histories generated by the model itself. ActionSplice corrects the active solver state when an action update arrives during sampling.
%ActionSplice addresses action updates received while the active chunk is still being sampled. It corrects the interrupted representation and resumes the remaining solver evaluations.

\par\noindent\textbf{Interactive condition changes and memory.}
Prior work addresses responsiveness to changing conditions through training objectives and memory management during inference. Delta Forcing~\citep{wu2026delta} constrains teacher supervision within an adaptive trust region during training to balance event responsiveness and temporal consistency. Echo-Forcing~\citep{wu2026echo} separates stable, recent, and recalled memory to support prompt switching and scene recall. LongLive~\citep{yang2025longlive} refreshes cached states after prompt switches, while Anchor Forcing~\citep{yang2026anchor} reconstructs the cache from anchor memories. Visko Orbis~\citep{gao2026visko} supports live prompt updates during streaming video generation. A revised action may be supplied by a user or downstream planner~\citep{tao2026navidrivevlm}. ActionSplice focuses on camera or action updates received while the active chunk is being sampled. CST estimates the counterfactual solver state that matched rollback would reach at the same solver step.

\par\noindent\textbf{Inference-time reuse and acceleration.}
Video diffusion methods reduce inference cost through caching and sparse computation. TeaCache~\citep{liu2025timestep} and FasterCache~\citep{lyu2025fastercache} reuse features across denoising steps, while Pyramid Attention Broadcast~\citep{zhao2025real} reuses attention outputs. Sparse VideoGen~\citep{xi2025sparse} and SparsePR~\citep{taghavi2026partition} reduce attention cost through spatiotemporal sparsity. Light Interaction~\citep{lu2026light} combines context selection, denoising reuse, and sparse attention for interactive generation. X-Cache~\citep{zeng2026x} and C$^3$ache~\citep{zhao2026c} reuse computation across chunks. Chorus~\citep{liu2026beyond} reuses intermediate information across similar requests while maintaining alignment with their prompts. ActionSplice corrects the active solver state toward its matched counterfactual at the interruption step, then resumes sampling without replaying completed evaluations.

\par\noindent\textbf{Revisable denoising and editing of intermediate states.}
Diffusion Forcing~\citep{chen2024diffusion} assigns independent noise levels to tokens, while Diffusion ReRoll~\citep{kim2026diffusion} selectively re-noises stable regions to enable revision across a temporal horizon. For source editing, SDEdit~\citep{meng2021sdedit} uses re-noising, EDICT~\citep{wallace2023edict} uses inversion, and Layered Diffusion Brushes~\citep{gholami2025streamlining} caches latents for localized edits. FateZero~\citep{qi2023fatezero} reuses attention maps from video inversion to preserve structure and temporal consistency. ActionSplice operates on an unfinished action-conditioned rollout. CST learns from matched rollback pairs to approximate the solver state induced by the requested intervention at the current step. Sampling resumes from the corrected state without restarting or replaying completed evaluations.

\begin{figure*}[t]
    \centering
    \includegraphics[width=\textwidth]{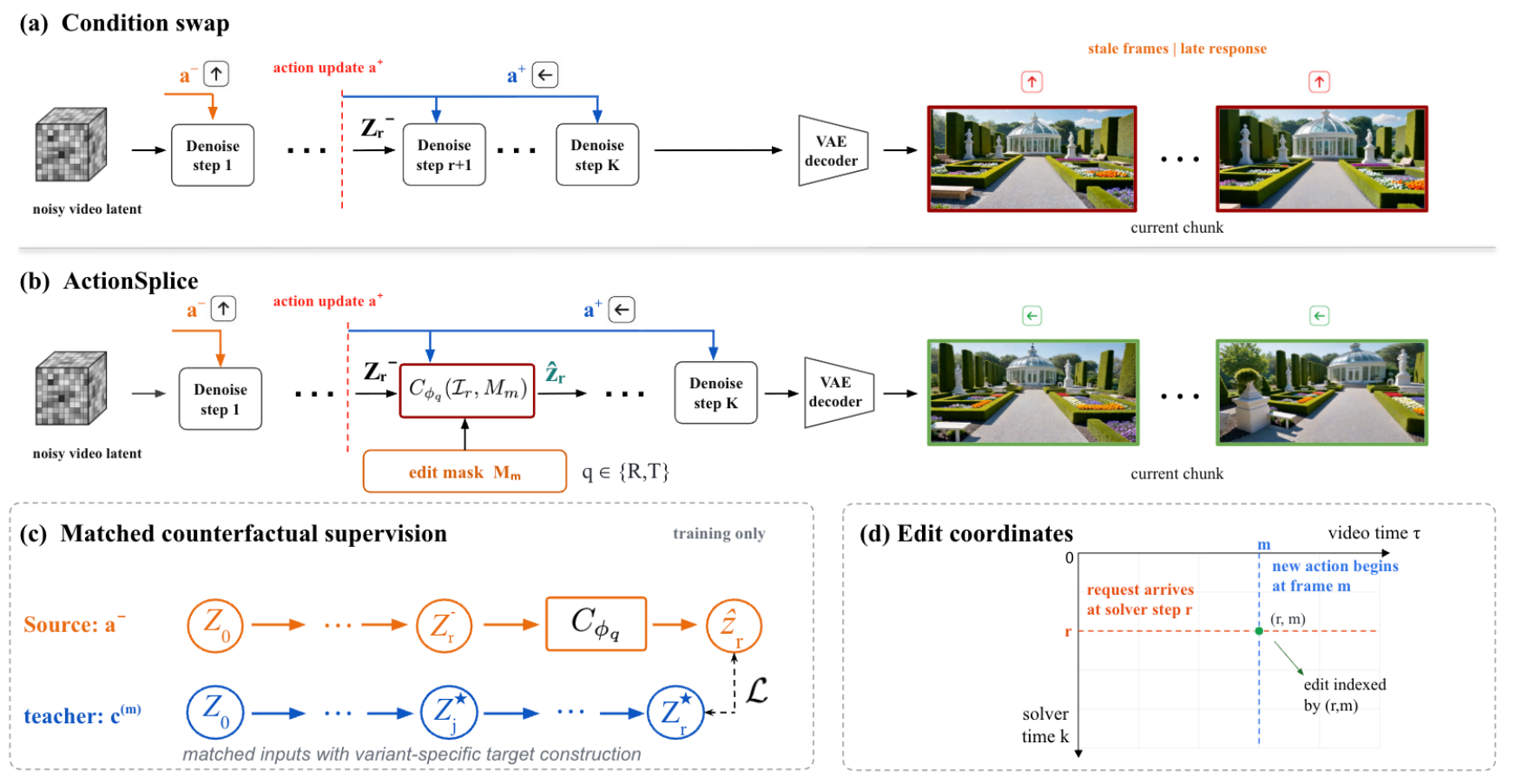}
    \caption{ActionSplice overview. (a) Condition swapping retains the interrupted representation. (b) CST predicts a masked correction at receipt step \(r\) and resumes sampling. (c) Matched rollback supplies training targets, with prefix clamping for \(\mathrm{CST}_{T}\). (d) The receipt step \(r\) indexes solver time and the boundary \(m\) indexes latent video time.}
    \label{fig:cst_method_overview}
\end{figure*}
\section{Method}
\label{sec:method}

ActionSplice corrects the interrupted solver state toward its matched
counterfactual. The frozen world model then resumes sampling without
replaying completed evaluations (Figure~\ref{fig:cst_method_overview}).

\subsection{Problem Formulation}
\label{sec:problem}

A frozen world model generates a chunk with \(T\) temporal latent positions
through \(K\) solver evaluations, initially conditioned on action \(a^-\).
An update arrives after \(r\in\{1,\ldots,K-1\}\) evaluations, requesting a
revised action \(a^+\). Previously completed chunks form the committed
history and remain unchanged. Let \(z_r^-\) denote a resumable representation
of the interrupted solver state at step \(r\).

The request specifies a boundary \(m\in\{0,\ldots,T-1\}\), the first latent
position assigned the revised action. For each position
\(i\), the requested action assignment \(c^{(m)}\) and
correction mask \(M_m\) are
\begin{equation}
c_i^{(m)}=
\begin{cases}
a^-, & i<m,\\
a^+, & i\geq m,
\end{cases}
\qquad
M_m[i]=\mathbf{1}[i\geq m].
\label{eq:cst_control_mask}
\end{equation}
The receipt step \(r\) specifies when sampling is interrupted, while \(m\)
specifies where the requested action changes within the chunk.
\(\mathrm{CST}_{R}\) retargets the entire chunk with \(m=0\).
\(\mathrm{CST}_{T}\) requests the revised action in the suffix with
\(0<m<T\), retaining the previous action assignment in the prefix. We
incorporate the request into the active chunk without replaying completed
evaluations.

\subsection{Matched Counterfactual Targets}
\label{sec:cst_supervision}

Let \(z_r^\star\) denote the matched counterfactual state representation
after evaluation \(r\). To construct it, the source and target branches
share the prompt, initial state, committed history, solver schedule, and
stochastic inputs when applicable. The source branch runs the first \(r\)
evaluations under \(a^-\). For \(\mathrm{CST}_{R}\), the target branch runs
the same evaluations under \(a^+\) to obtain \(z_r^\star\).

For \(\mathrm{CST}_{T}\), the target branch runs under the mixed action
assignment \(c^{(m)}\) while enforcing a prefix constraint after each
evaluation. Target prefix quantities are replaced by the corresponding
source quantities from the same solver step, while the suffix retains the
target branch values. Consequently,
\[
z_r^\star[i]=z_r^-[i],\qquad i<m.
\]
Appendix~\ref{app:transport_details} specifies the clamped quantities,
conditioning, and operation order for each implementation. Rollback is used
only to construct training targets and evaluation references, not during
ActionSplice inference.

\subsection{Counterfactual State Transport}
\label{sec:cst_transport}

The corrector input \(\mathcal I_r\) contains \(z_r^-\), the initial state,
preceding chunk, old and revised action conditioning, receipt step, and
retained sampling information. Appendix~\ref{app:transport_details}
specifies the input encoding. For edit type \(q\in\{R,T\}\), the corrector
\(C_{\phi_q}\) with learned parameters \(\phi_q\) predicts a residual that is restricted by the temporal mask:

\begin{equation}
\widehat z_r
=
z_r^-+M_m\odot C_{\phi_q}(\mathcal I_r,M_m),
\label{eq:cst_masked_transport}
\end{equation}
where \(\odot\) denotes elementwise multiplication. The mask broadcasts over channel and spatial dimensions, ensuring
\(
(\mathbf 1-M_m)\odot(\widehat z_r-z_r^-)=0.
\)
For \(\mathrm{CST}_{T}\), this preserves prefix coordinates at the
intervention step but does not guarantee an identical decoded prefix after
further sampling.

Sampling resumes from \(\widehat z_r\) under conditioning constructed from
\(c^{(m)}\). Appendix~\ref{app:transport_details} specifies scheduler
conversion, prefix handling, and cache operations. Each
intervention uses one corrector call and the remaining \(K-r\) solver
evaluations, whereas full rollback performs \(K\) solver evaluations after receipt. The corrector, state conversion, and cache updates add overhead
beyond this count.

\subsection{Corrector Training}
\label{sec:cst_training}

We train a separate corrector for each backbone and intervention type. Each
corrector is a six-block residual 3D encoder-decoder conditioned on action
inputs and receipt step through FiLM~\citep{perez2018film}. We optimize only
the corrector parameters \(\phi_q\). The world model and sampler remain
frozen, and the mask \(M_m\) is fixed by the supplied boundary \(m\). We minimize masked normalized mean squared error:
\begin{equation}
\mathcal L=
\frac{\|M_m\odot(\widehat z_r-z_r^\star)\|_2^2/N_m}
{\max\!\left(\|M_m\odot z_r^\star\|_2^2/N_m,\epsilon\right)},
\label{eq:cst_matching}
\end{equation}
where \(N_m\) counts editable scalar elements after broadcasting the mask
and \(\epsilon=10^{-8}\). Appendix~\ref{app:transport_details} specifies
the input encoding and corrector architecture.

Training pairs are collected from recurrent rollouts whose committed
histories include outputs from earlier corrections. At each interruption,
the source and target branches use the same committed history. Training
uses stored capture pairs, so gradients do not propagate through the rollout that produced this history. Appendix~\ref{app:optimization}
details initialization and checkpoint selection.

\section{Experiments}

\subsection{Experimental Setup and Evaluation Protocol} 
\label{sec:experimental_setup}
We evaluate ActionSplice on minWM--Wan Action2V~\citep{zhao2026minwm} and HY-WM1.5~\citep{hunyuanworld2025hy}, using four solver evaluations per chunk. We train a separate corrector for each backbone and CST variant. Both backbones share a prompt-disjoint split of 120 training, 30 validation, and 30 test scenes. Training captures come from recurrent rollouts whose committed histories include earlier corrections. Checkpoint selection uses the validation set. The main evaluation uses receipt step $r=2$, with 30 matched groups of test prompts for $\mathrm{CST}_{R}$ and 90 matched combinations of test prompts and temporal boundaries for $\mathrm{CST}_{T}$, covering $m\in\{1,2,3\}$ for each prompt. Transitions cover all six directed pairs among forward, backward, and yaw left motion. Additional experiments vary the receipt step, action transition, temporal boundary, and number of interruptions.

We compare CST with waiting, condition swapping, partial rollback, re-noising, and matched rollback. Waiting applies the revised action in the next chunk. Condition swapping changes the conditioning for the remaining evaluations without correcting the interrupted state representation. Partial rollback restores the preceding solver checkpoint and repeats one completed evaluation. Re-noising perturbs a clean estimate to the noise level preceding the final two evaluations and runs them under the revised action. The reference is full rollback for $\mathrm{CST}_{R}$ and rollback with prefix clamping for $\mathrm{CST}_{T}$. Methods within each group share the prompt, initial state, committed history, action update, receipt step, and solver schedule. Stochastic inputs are matched where applicable, with an additional fixed Gaussian draw for re-noising.

We measure reference fidelity using LPIPS and PSNR over the active chunk for $\mathrm{CST}_{R}$ and its editable suffix for $\mathrm{CST}_{T}$. Prefix LPIPS compares the retained region with the matched clamped reference. Boundary error measures the discrepancy between generated and reference frame changes at the start of the editable region. Camera trajectory and HY-WorldPlay benchmark results use arithmetic means under their respective protocols. Blinded human annotations measure action following as the percentage of examples showing the revised action in the editable region and stale frames as the mean number of frames before its first visible response. Pixel-ready latency starts at action receipt and ends when the decoded active chunk is available for immediate intervention methods or when the decoded next chunk is available for waiting. We compute speedup as the ratio of waiting's median latency to the method's median latency. Timing uses an NVIDIA A100 for minWM and an NVIDIA H100 for HY-WM1.5, with comparisons restricted to the same backbone. Appendix~\ref{app:evaluation_protocol} details capture grouping, metric computation, and timing.

\begin{figure}[!b]
    \centering
    \includegraphics[width=0.99\linewidth]{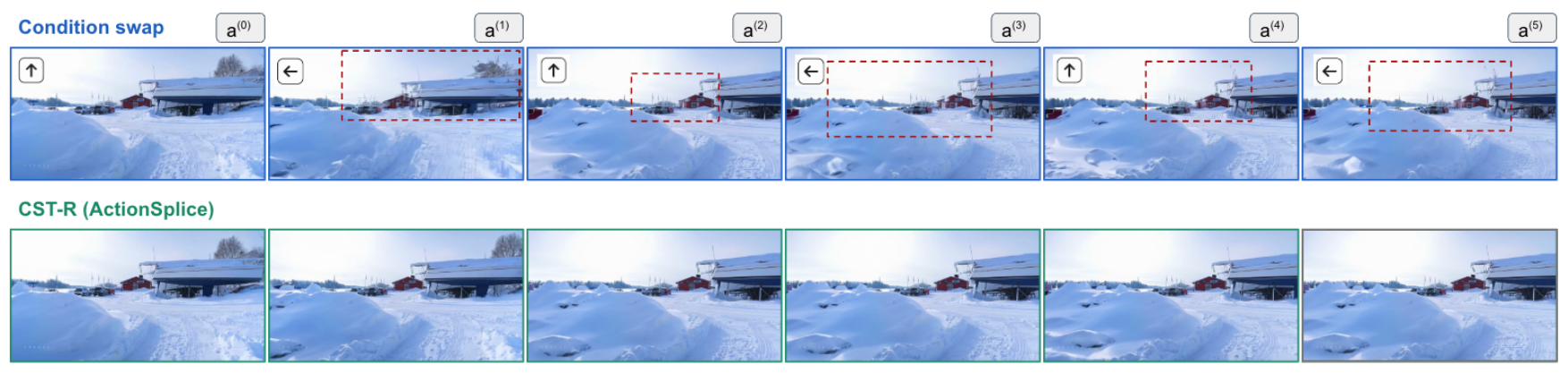}
    \caption{Qualitative HY-WM1.5 rollout with five repeated action updates
across six action segments. Each update arrives at \(r=2\) of \(K=4\).
Condition swapping (top) retains stale motion after the updates, whereas \(\mathrm{CST}_{R}\) (bottom) follows each revised
action within the active chunk.}
    \label{fig:recurrent_example}
\end{figure}
\subsection{Whole-Chunk Action Updates}
\label{sec:results_cst_r}

We evaluate whole-chunk retargeting with an action update received after
$r=2$ of the $K=4$ solver evaluations. The revised action is intended to control the entire active chunk. Table~\ref{tab:model_benchmarks} compares methods under the matched protocol, with full rollback as the reference. Appendices~\ref{app:interrupt_sweeps} and~\ref{app:transition_breakdown} report sweeps over receipt steps and results for all six directed action transitions.

\begin{table*}[t]
    \centering
    \caption{Whole-chunk retargeting at $r=2$ of $K=4$ solver evaluations over 30 test prompts. Fidelity is measured over the active chunk against matched full rollback. Bold and underlined values mark the best and second-best non-oracle results within each backbone.}
    \label{tab:model_benchmarks}
    \resizebox{\textwidth}{!}{%
    \begin{tabular}{llccccc}
        \toprule
        Model & Method &
        \makecell{Pixel-ready\\latency (ms) $\downarrow$} &
        \makecell{Speedup\\$\uparrow$} &
        LPIPS $\downarrow$ &
        PSNR $\uparrow$ &
        \makecell{Boundary\\error $\downarrow$} \\
        \midrule
        \multirow{6}{*}{minWM} & Wait
            & 6815.0 & $1.00\times$
            & 0.3159 & 12.95 & 0.0215 \\
        & Condition Swap
            & \textbf{2498.1}
            & $\mathbf{2.73}\times$ & 0.3155 & \underline{12.97} & 0.0215 \\
        & Partial Rollback-$d=1$
            & 2979.3 & $2.29\times$
            & \underline{0.3111} & 12.95 & \underline{0.0212} \\
        & Re-noising (rerun 2)
            & 2556.4
            & $2.67\times$ & 0.3160 & 12.92 & 0.0214 \\
        & Full Rollback
            & 3481.4 & $1.96\times$
            & -- & -- & -- \\
        \rowcolor{blueLight}
        & $\mathbf{CST}_{R}$
            & $\underline{2537.1}$
            & $\underline{2.69}\times$ & \textbf{0.1214}
            & \textbf{16.30} & \textbf{0.0175} \\
        \midrule
        \multirow{6}{*}{HY-WM1.5} & Wait
            & 11711.2 & $1.00\times$
            & 0.3682 & 14.94 & 0.0379 \\
        & Condition Swap
            & \textbf{7052.1}
            & $\mathbf{1.66}\times$
            & 0.3446 & 15.20 & \underline{0.0368} \\
        & Partial Rollback-$d=1$
            & 7942.8 & $1.47\times$
            & \underline{0.1674} & \underline{18.89} & 0.0399 \\
        & Re-noising
            & \underline{7101.2}
            & $\underline{1.65}\times$
            & 0.3480 & 15.08 & 0.0379 \\
        & Full Rollback (oracle)
            & 8706.8 & $1.35\times$
            & -- & -- & -- \\
        \rowcolor{blueLight}
        & $\mathbf{CST}_{R}$
            & 7124.0
            & $1.64\times$
            & \textbf{0.0831} & \textbf{20.51} & \textbf{0.0196} \\
        \bottomrule
    \end{tabular}%
    }
\end{table*}

$\mathrm{CST}_{R}$ achieves the lowest LPIPS and boundary error and the highest PSNR among non-oracle methods on both backbones. Compared with condition swapping, it reduces LPIPS by 61.5\% on minWM and 75.9\% on HY-WM1.5, with PSNR gains of 3.33 and 5.31\,dB, respectively.
On HY-WM1.5, CST also lowers LPIPS from 0.1674 to 0.0831 relative to partial rollback while reducing latency from 7942.8 to 7124.0\,ms. It achieves pixel-ready speedups of $2.69\times$ and $1.64\times$ over waiting on minWM and HY-WM1.5, respectively. These results show lower error against matched rollback without replaying completed solver evaluations.
%\FloatBarrier

\subsection{Within-Chunk Action Transitions}
\label{sec:results_cst_t}

We evaluate action transitions at boundary $m$ within the active chunk. Positions before $m$ should retain the previous action, while those from $m$ onward should follow the revised action. $\mathrm{CST}_{T}$ restricts its learned correction to this suffix. Table~\ref{tab:within_chunk_benchmarks} reports aggregate results against rollback with prefix clamping, and Appendix~\ref{app:boundary_sweep} provides results for each boundary.

\FloatBarrier

\begin{table*}[!htbp]
    \centering
    \caption{Action transitions within a chunk at $r=2$, $K=4$, over 30 prompts and boundaries $m\in\{1,2,3\}$ (90 groups). Fidelity uses matched rollback with prefix clamping. Bold and underlined values mark the best and second best results excluding the oracle for each backbone.}
    \label{tab:within_chunk_benchmarks}
    \resizebox{\textwidth}{!}{%
    \begin{tabular}{llcccc}
        \toprule
        Model & Method &
        \makecell{Prefix\\LPIPS $\downarrow$} &
        \makecell{Suffix\\LPIPS $\downarrow$} &
        \makecell{Suffix\\PSNR $\uparrow$} &
        \makecell{Boundary\\error $\downarrow$} \\
        \midrule
        \multirow{6}{*}{minWM} & Wait
            & 0.0863 & 0.2874 & 13.15 & 0.0552 \\
        & Condition Swap
            & 0.0867 & 0.2868 & 13.16 & 0.0548 \\
        & Partial Rollback-$d=1$
            & \underline{0.0842} & \underline{0.2773}
            & \underline{13.27} & \underline{0.0542} \\
        & Re-noising (rerun 2)
            & 0.0897 & 0.2889 & 13.14 & 0.0549 \\
        & Full Rollback (oracle)
            & -- & -- & -- & -- \\
        \rowcolor{blueLight}
        & $\mathbf{CST}_{T}$
            & \textbf{0.0740} & \textbf{0.1258} & \textbf{16.46}
            & \textbf{0.0511} \\
        \midrule
        \multirow{6}{*}{HY-WM1.5} & Wait
            & 0.0715 & 0.3612 & 15.06 & 0.0497 \\
        & Condition Swap
            & 0.0763 & 0.2731 & 15.53 & 0.0451 \\
        & Partial Rollback-$d=1$
            & \underline{0.0498} & \underline{0.0809} & \underline{21.52}
            & \underline{0.0269} \\
        & Re-noising (rerun 2)
            & 0.0933 & 0.2790 & 15.51 & 0.0464 \\
        & Full Rollback (oracle)
            & -- & -- & -- & -- \\
        \rowcolor{blueLight}
        & $\mathbf{CST}_{T}$
            & \textbf{0.0412} & \textbf{0.0615}
            & \textbf{25.05} & \textbf{0.0215} \\
        \bottomrule
    \end{tabular}%
    }
\end{table*}

\begin{table*}[t]
    \centering
    \caption{Human evaluation of action responsiveness on HY-WM1.5 over 30 matched \(\mathrm{CST}_R\) groups and 90 matched \(\mathrm{CST}_T\) groups. Action following measures responses within the editable region; mean stale frames measures the delay to the first visible response. }
    \label{tab:human_action_response}
    \resizebox{0.82\textwidth}{!}{%
    \begin{tabular}{llcccc}
        \toprule
        Model & Method &
        \multicolumn{2}{c}{$\mathrm{CST}_{R}$} &
        \multicolumn{2}{c}{$\mathrm{CST}_{T}$} \\
        \cmidrule(lr){3-4}\cmidrule(lr){5-6}
        & &
        \makecell{Action\\following (\%) $\uparrow$} &
        \makecell{Mean stale\\frames $\downarrow$} &
        \makecell{Action\\following (\%) $\uparrow$} &
        \makecell{Mean stale\\frames $\downarrow$} \\
        \midrule
        \multirow{6}{*}{HY-WM1.5} & Wait
            & 20.0 & 15.37 & 26.7 & 7.53 \\
        & Condition Swap
            & 83.3 & 8.90 & 86.7 & 3.73 \\
        & Partial Rollback-$d=1$
            & \underline{86.7} & \underline{3.00}
            & \underline{93.3} & \underline{0.93} \\
        & Re-noising
            & 76.7 & 9.90 & 86.7 & 4.40 \\
        & Full Rollback (oracle)
            & 100.0 & 0.00 & 100.0 & 0.00 \\
        \rowcolor{blueLight}
        & $\mathbf{CST}$
            & \textbf{96.7} & \textbf{1.07}
            & \textbf{100.0} & \textbf{0.40} \\
        \bottomrule
    \end{tabular}%
    }
\end{table*}

$\mathrm{CST}_{T}$ achieves the lowest prefix LPIPS, suffix LPIPS, and boundary error and the highest suffix PSNR among non-oracle methods on both backbones. Compared with condition swapping, it reduces suffix LPIPS by 56.1\% on minWM and 77.5\% on HY-WM1.5, with suffix PSNR gains of 3.30 and 9.52\,dB, respectively. On HY-WM1.5, CST also lowers suffix LPIPS from 0.0809 to 0.0615 relative to partial rollback. Lower prefix LPIPS and boundary error indicate closer agreement with the clamped reference before and across the action boundary.

Table~\ref{tab:human_action_response} reports blinded human evaluation on HY-WM1.5. $\mathrm{CST}_R$ follows the revised action in 96.7\% of cases with 1.07 mean stale frames, compared with 83.3\% and 8.90 frames for condition swapping. For $\mathrm{CST}_T$, action-following reaches 100.0\% with 0.40 mean stale frames, compared with 86.7\% and 3.73 frames for condition swapping.

\subsection{Comparison with Prior Work}
\label{sec:prior_comparison}

Table~\ref{tab:hyworldplay_quality} reports a secondary evaluation of
$\mathrm{CST}_{R}$ under the HY-WorldPlay protocol for 200 prompts used by
\citet{lu2026light}. Using the reproduced HY-WM1.5 backbone, CST achieves
the highest reported PSNR and SSIM and the lowest LPIPS in both comparisons.
This evaluation complements the main experiments on in-flight action-update
fidelity and latency.

\begin{table}[!htbp]
    \centering
    \caption{Global video quality under the HY-WorldPlay protocol. $\mathrm{CST}_{R}$ uses reproduced HY-WM1.5. Bold and underlined values mark the best and second-best reported results.}
    \label{tab:hyworldplay_quality}
    \resizebox{\linewidth}{!}{%
    \begin{tabular}{llcccccc}
        \toprule
        Model & Method &
        \multicolumn{3}{c}{vs. Original} &
        \multicolumn{3}{c}{Self-Comparison} \\
        \cmidrule(lr){3-5}\cmidrule(lr){6-8}
        & & PSNR $\uparrow$ & SSIM $\uparrow$ & LPIPS $\downarrow$
          & PSNR $\uparrow$ & SSIM $\uparrow$ & LPIPS $\downarrow$ \\
        \midrule
        \multirow{6}{*}{HY-WorldPlay}
          & Original~\citep{hunyuanworld2025hy} & -- & -- & -- & 18.60 & 0.5678 & 0.2051 \\
          & SVG~\citep{xi2025sparse} & 19.48 & 0.6028 & 0.2209 & 17.75 & 0.5299 & 0.2187 \\
          & BSA~\citep{longcat2025video} & 15.94 & 0.4639 & 0.3755 & 15.44 & 0.4205 & 0.3720 \\
          & TeaCache~\citep{liu2025timestep} & 20.90 & \underline{0.6588}
              & 0.1892 & \underline{18.86} & 0.5743 & 0.2054 \\
          & Light Interaction~\citep{lu2026light} & \underline{24.81} & 0.6500
              & \underline{0.1788} & 18.85 & \underline{0.5854}
              & \underline{0.1963} \\
          \rowcolor{blueLight}
          & $\mathrm{CST}_{R}$ & \textbf{25.66} & \textbf{0.6902}
              & \textbf{0.1337} & \textbf{20.38} & \textbf{0.6267} & \textbf{0.1587} \\
        \bottomrule
    \end{tabular}
    }
\end{table}
%$\mathrm{CST}_{R}$ obtains the best reported PSNR, SSIM, and LPIPS in both comparisons, indicating that in-flight state transport preserves global video quality under the established benchmark.

\subsection{Camera-Action Following}
\label{sec:camera_action_following}

We evaluate camera-action following with the rotational and translational
trajectory errors $R_{\mathrm{dist}}$ and $T_{\mathrm{dist}}$ from
\citet{hunyuanworld2025hy}. We estimate camera poses with ViPE~\citep{huang2025vipe} and follow
WorldPlay by expressing estimated and target trajectories relative to
their first poses and rescaling estimated translations using the furthest target frame. Table~\ref{tab:camera_action_following} reports errors over the active chunk for $\mathrm{CST}_{R}$ and the editable suffix for $\mathrm{CST}_{T}$. Published results evaluate videos of 61 frames and provide context, although the different evaluation intervals prevent a direct comparison.

%Published results use 61-frame videos and provide a reference for camera-control performance, although the different evaluation intervals prevent a direct comparison.

\begin{table}[!htbp]
    \centering
    \caption{Camera action following at $r=2$ over 30 prompts, averaging three boundaries for $\mathrm{CST}_{T}$. Published results provide context under different evaluation intervals. Lower is better.}
    \label{tab:camera_action_following}
    \small
    \setlength{\tabcolsep}{4pt}
    \begin{tabular}{lcc}
        \toprule
        Method &
        $R_{\mathrm{dist}}\downarrow$ &
        $T_{\mathrm{dist}}\downarrow$ \\
        \midrule
        %\multicolumn{3}{l}{} \\
        CameraCtrl~\citep{he2024cameractrl}      & 0.037 & 0.341 \\
        %Gen3C            & \textbf{0.024} & 0.477 \\
        VMem~\citep{li2025vmem}            & 0.048 & 0.219 \\
        Matrix-Game 2.0~\citep{he2025matrix} & 0.287 & 0.843 \\
        WorldPlay~\citep{hunyuanworld2025hy} & 0.031 & 0.121 \\
        \midrule
        %\multicolumn{3}{l}{} \\
        \rowcolor{blueLight}
        $\mathrm{CST}_{R}$ & 0.048 & 0.025 \\
        \rowcolor{blueLight}
        $\mathrm{CST}_{T}$ & 0.053 & 0.054 \\
        \bottomrule
    \end{tabular}
\end{table}

\subsection{Repeated In-Flight Action Updates}
\label{sec:repeated_updates}

\begin{figure*}[t]
    \centering
    \includegraphics[width=\textwidth]
{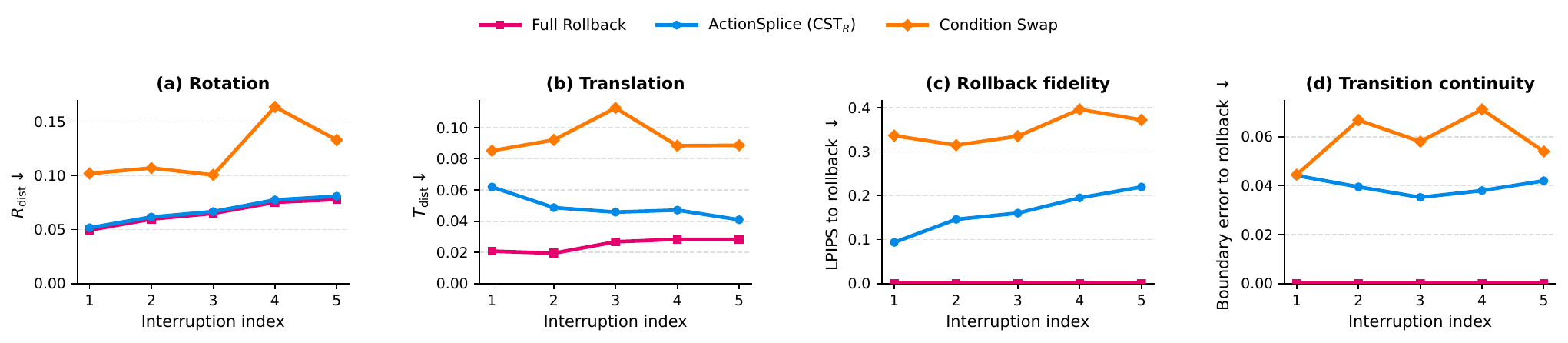}
    \caption{Repeated in-flight updates on HY-WM1.5 at $r=2$ of $K=4$.
The $\mathrm{CST}_{R}$ corrector is trained on trajectories with three
interruptions and tested on five without further training. Curves show
mean errors across test prompts. Camera errors are measured against
the requested trajectory, while LPIPS and boundary error are measured
against sequential full rollback.}
    \label{fig:repeated_updates}
\end{figure*}

We evaluate error accumulation across five action updates and six segments
on HY-WM1.5. Each update arrives at $r=2$ of $K=4$ solver evaluations.
The $\mathrm{CST}_{R}$ corrector is trained on trajectories with three
interruptions and evaluated on five-interruption sequences without further
training. CST, condition
swapping, and sequential full rollback use matched prompts, action
schedules, initial states, and stochastic inputs.
Figure~\ref{fig:repeated_updates} reports camera-trajectory errors
against the requested motion and LPIPS and boundary error against
sequential full rollback. Across all five interruptions, $\mathrm{CST}_{R}$ has lower errors than condition swapping. Its rollback-relative LPIPS increases with interruption index, showing growing deviation from
the reference. Figure~\ref{fig:recurrent_example} shows a qualitative example of a rollout with six action segments.

\begin{figure*}[!b]
    \centering
    \includegraphics[width=\textwidth]{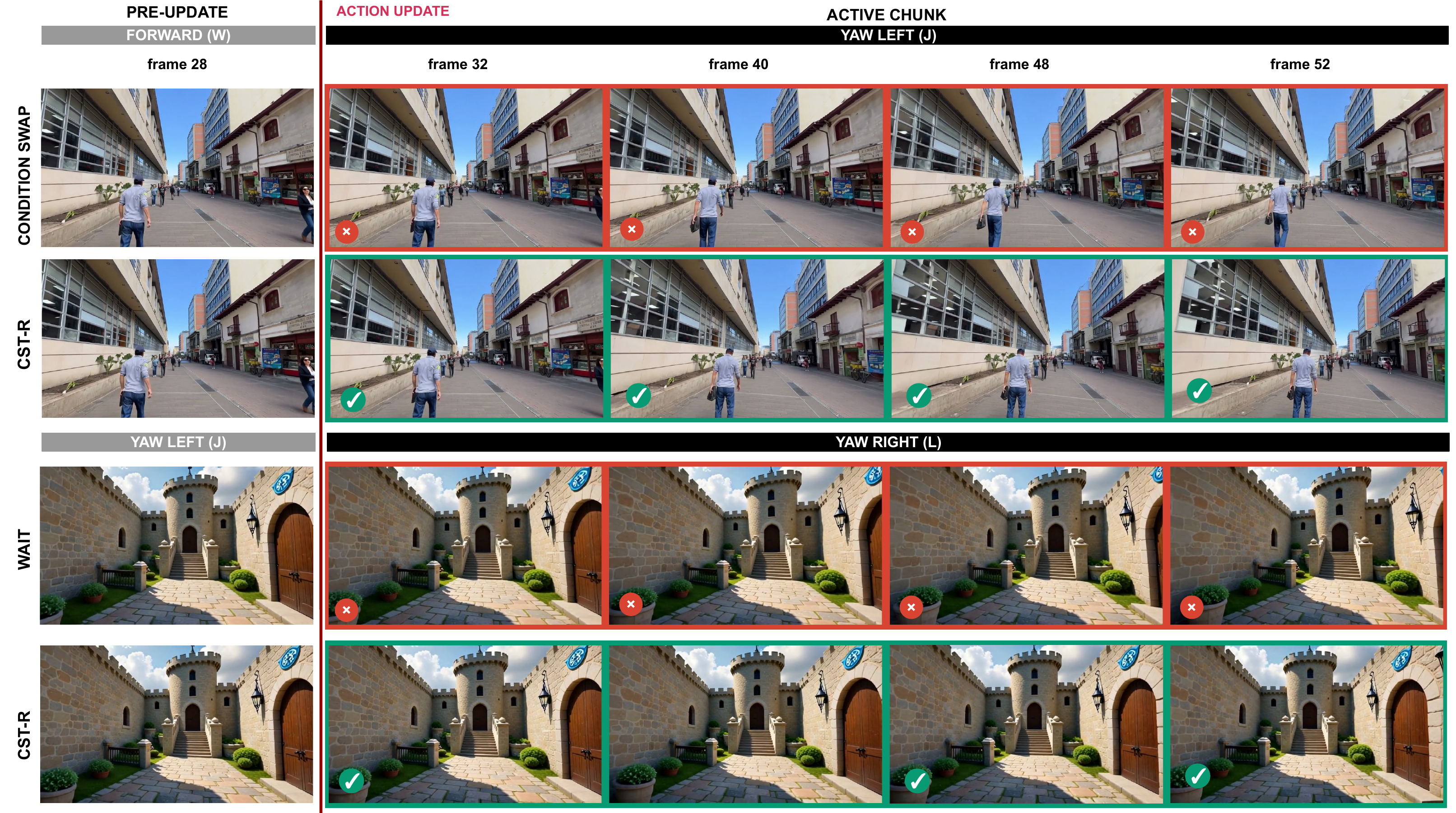}
    \caption{Whole-chunk action-update examples. Top pair shows a
forward to yaw-left request on HY-WM1.5, where condition swapping
continues forward motion. Bottom pair shows yaw-left to yaw-right request on minWM, where waiting delays the update until next chunk.
$\mathrm{CST}_{R}$ responds within the active chunk in both examples. Red and green borders mark stale and responsive frames.}
    \label{fig:qualitative_comparison}
\end{figure*}

\subsection{Qualitative Comparisons}
\label{sec:qualitative_comparisons}

Figure~\ref{fig:qualitative_comparison} shows whole-chunk updates on
both backbones. On HY-WM1.5, condition swapping continues forward
after a yaw-left request, while $\mathrm{CST}_{R}$ turns within the
active chunk. On minWM, waiting delays a yaw-right request until the
next chunk, while $\mathrm{CST}_{R}$ responds within the active chunk.
In both examples, CST makes the revised action visible without
restarting sampling.

\begin{figure*}[t]
    \centering
    \includegraphics[width=0.8\textwidth]{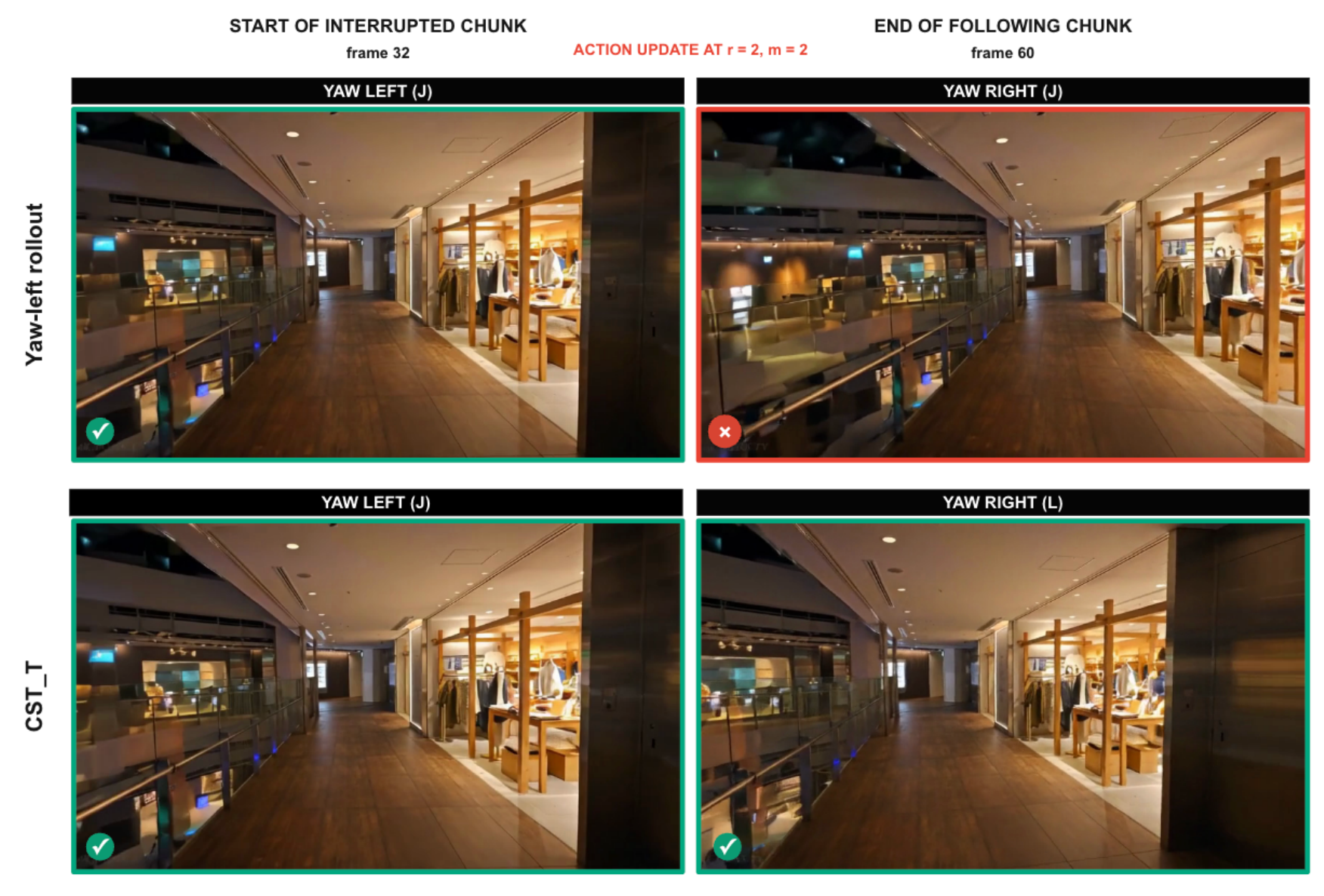}
    \caption{Within-chunk action transition requested at $r=2$ with boundary
$m=2$. Top rollout continues yaw left, while $\mathrm{CST}_{T}$
follows the requested change from yaw left to yaw right. Columns show
the start of the interrupted chunk and the end of the following chunk.
Green and red borders indicate agreement and disagreement with the
requested action, respectively.}
    \label{fig:qualitative_cst_t}
\end{figure*}

Figure~\ref{fig:qualitative_cst_t} shows a within-chunk transition
requested at $r=2$, with a change from yaw left to yaw right at $m=2$.
The comparison rollout follows yaw left throughout the chunk.
$\mathrm{CST}_{T}$ retains yaw-left motion in the prefix and follows
yaw right in the suffix. Both rollouts continue through the next chunk to show how their trajectories develop.

%\(\mathcal{L}\)
%\(C_{\phi_q}\)
%\subsection{Computation accounting}

\subsection{Ablations}
\label{sec:ablations}

We ablate matched counterfactual supervision and temporal masking
under the full model's evaluation protocol. Removing the mask allows
$\mathrm{CST}_{T}$ to modify prefix coordinates and tests the effect
of restricting corrections to the suffix. The supervision ablation pairs each source with
a target from a different prompt at the same solver step.
Table~\ref{tab:actionsplice_ablations} reports action error, rollback-relative
fidelity, boundary error, and prefix LPIPS. Action error averages the sum of rotation and translation errors
between estimated and requested relative camera motions, normalized by their respective command magnitudes. Appendix~\ref{app:ablation_details}
provides the formula and ablation settings.

\begin{center}
\begin{minipage}{\linewidth}
    \centering
    \captionof{table}{ActionSplice ablations on HY-WM1.5. RB-LPIPS compares the
active chunk with matched full rollback for $\mathrm{CST}_{R}$ and
the editable suffix with matched prefix-clamped rollback for
$\mathrm{CST}_{T}$. Prefix LPIPS uses the clamped reference. Lower is better.}
    \label{tab:actionsplice_ablations}
    \small
    \setlength{\tabcolsep}{4.5pt}
    \renewcommand{\arraystretch}{1.08}
    \resizebox{\linewidth}{!}{%
    \begin{tabular}{lccccccc}
        \toprule
        & \multicolumn{3}{c}{$\mathrm{CST}_{R}$}
        & \multicolumn{4}{c}{$\mathrm{CST}_{T}$} \\
        \cmidrule(lr){2-4}
        \cmidrule(lr){5-8}
        Configuration
        & \makecell{Action\\error $\downarrow$}
        & \makecell{RB-LPIPS\\$\downarrow$}
        & \makecell{Boundary\\error $\downarrow$}
        & \makecell{Action\\error $\downarrow$}
        & \makecell{RB-LPIPS\\$\downarrow$}
        & \makecell{Boundary\\error $\downarrow$}
        & \makecell{Prefix\\LPIPS $\downarrow$} \\
        \midrule
        \rowcolor{blueLight}
        Full model
            & \textbf{0.6192} & \textbf{0.0831} & \textbf{0.0196}
            & \textbf{0.3283} & \textbf{0.0615} & \textbf{0.0215} & \textbf{0.0412} \\
        \midrule
        Without hard mask
            & -- & -- & --
            & 0.4092 & 0.0787 & 0.0240 & 0.0604 \\
        Unmatched supervision
            & 4.6889 & 0.7797 & 0.0435
            & 3.0415 & 0.7719 & 0.0349 & 0.4209 \\
        \bottomrule
    \end{tabular}%
    }
\end{minipage}
\end{center}

The full model achieves the lowest error on every reported metric.
Without the hard mask, $\mathrm{CST}_{T}$ prefix LPIPS rises from
0.0412 to 0.0604 and boundary error from 0.0215 to 0.0240.
Unmatched supervision worsens every metric, supporting the use
of matched source-target pairs.

\section{Conclusion}

We introduced ActionSplice for incorporating action changes into an
active chunk during sampling. A learned corrector moves the interrupted
state representation toward its matched counterfactual.
\(\mathrm{CST}_{R}\) retargets the active chunk, while
\(\mathrm{CST}_{T}\) restricts the correction to a temporal suffix.
Across minWM and HY-WM1.5, CST improves fidelity to matched rollback
relative to condition swapping. $\mathrm{CST}_{R}$ also reduces pixel-ready
latency relative to waiting and maintains lower errors than condition swapping
across five repeated updates. CST enables sampling to resume under
revised actions without repeating completed solver evaluations.

\bibliography{iclr2027_conference}
\bibliographystyle{iclr2027_conference}

\section*{Supplementary Material}
\appendix
\section{Implementation and Training Details}

\subsection{Backbone-Specific Transport and Resumption}
\label{app:transport_details}

minWM receives camera view matrices and intrinsics, while HY-WM1.5 also
receives discrete action labels. For \(\mathrm{CST}_{T}\), the conditioning
switches action at latent position \(m\). Counterfactual target construction
first computes the source trajectory, then clamps the target prefix using
source quantities from the same solver step. The target is captured at
receipt step \(r\). This source trajectory is not used during inference.
During continuation, minWM restores the prefix cached at receipt, while
HY-WM1.5 continues without repeated prefix replacement.

The corrector receives the interrupted and initial states, the preceding
committed chunk, old and revised action conditioning, and the receipt step.
HY-WM1.5 continuation is deterministic and requires no noise trace. For
\(\mathrm{CST}_{T}\), the temporal mask is provided as an input channel and
also restricts the predicted residual. \(\mathrm{CST}_{R}\) additionally
conditions on the interruption index, while \(\mathrm{CST}_{T}\) does not.

Each corrector uses six FiLM residual blocks in a 3D encoder-decoder. Two
spatial downsampling stages use channel widths 128, 256, and 512, followed
by interpolation and skip additions. Each block combines a
\(1\times3\times3\) spatial convolution with a \(3\times1\times1\) temporal
convolution, GroupNorm, and SiLU. Camera conditioning contains 45 features
per latent position from old, revised, and relative poses and normalized
intrinsics. A two-layer MLP embeds each position. We pool the mean,
first-position, and last-position embeddings and combine them with receipt-step
conditioning and, for \(\mathrm{CST}_{R}\), an interruption-index embedding.
The conditioning width is 256. minWM and HY-WM1.5 use 16 and 32 latent
channels, respectively.

HY-WM1.5 rebuilds its context cache from committed frames before
continuation, while minWM reuses its existing caches.

\subsection{Dataset Construction}

Each prompt has one fixed seed. Captures follow the prompt-disjoint split
and action transitions described in the main text. Each capture stores source
and target solver traces from which training pairs are extracted at receipt
steps $r\in\{1,2,3\}$.

For $\mathrm{CST}_{R}$, each prompt produces one recurrent trajectory with
three interruption events and four action segments. Each action persists for
three or four chunks, which produces trajectories of 48 to 64 temporal latent positions.
The three events yield 540 captures in total, with 360 for training, 90 for
validation, and 90 for testing. For $\mathrm{CST}_{T}$, each prompt produces
one recurrent capture for each boundary $m\in\{1,2,3\}$. This gives the same 360/90/90
capture split. The six action transitions and three temporal boundaries are
balanced within each partition. Capture counts refer to interruption events,
not to the number of solver-step pairs extracted from their saved traces.
Later events use committed chunks generated after earlier corrections.
Source and target branches share this history at each event, and training
does not backpropagate through capture generation.

\subsection{Training Objective and Optimization}
\label{app:optimization}

We train each corrector for 4,000 steps with AdamW, batch size 1, weight decay
$10^{-4}$, and gradient clipping at 1.0. The learning rate follows a cosine
schedule from $10^{-5}$ to $10^{-6}$. Each backbone and variant has a separate corrector trained only with representation-matching NMSE while the world model and sampler remain frozen.

We average the normalized loss from Section~\ref{sec:cst_training} across examples. For \(\mathrm{CST}_{R}\), the loss covers the full state. For \(\mathrm{CST}_{T}\), it covers only the editable suffix. This loss mask is distinct from the mask applied to the predicted correction.

We maintain an exponential moving
average of the corrector weights with decay 0.999 and evaluate it every 200
steps on the validation set. We select the EMA checkpoint with the lowest
validation transport NMSE and use it for test evaluation.

\subsection{Evaluation Protocol}
\label{app:evaluation_protocol}

\paragraph{Comparison groups.}
We use the matched groups defined in the main text and aggregate each
backbone and variant separately. For $\mathrm{CST}_{R}$, we average the three
interruption-level measurements within each prompt, producing 30 prompt
groups from 90 test captures. Receipt-step sweeps vary $r$ within the same
trajectory. Each $\mathrm{CST}_{T}$ prompt-boundary pair is one group, giving
90 groups; the three boundaries for a prompt are not independent scenes.

\paragraph{Decoded evaluation intervals.}
For a rollout with $L$ latent positions and $F$ decoded frames, let $j$
be the active chunk's first latent index. Decoded indices are
\begin{equation}
 g(u)=\operatorname{round}\!\left(u\frac{F-1}{L-1}\right),
 \qquad b=g(j+m),\qquad e=\min\{F,g(j+T)\}.
\end{equation}
Rounding uses the nearest integer. The active chunk,
editable region, and prefix occupy $[g(j),e)$, $[b,e)$, and $[g(j),b)$,
respectively. The boundary $m$ indexes latent positions.

\paragraph{Reference fidelity.}
Let $x_t$ and $x_t^\star$ be generated and matched rollback frames with RGB
values in $[0,1]$. The reference follows Section~\ref{sec:cst_supervision},
including prefix clamping for $\mathrm{CST}_{T}$. For the editable decoded
interval $\mathcal E=[b,e)$, we average framewise AlexNet LPIPS and compute
PSNR from the region's mean squared error,
\begin{equation}
\operatorname{MSE}_{\mathcal E}
=\frac{1}{|\mathcal E|D}\sum_{t\in\mathcal E}\|x_t-x_t^\star\|_2^2,
\qquad
\operatorname{PSNR}_{\mathcal E}
=10\log_{10}\frac{1}{\operatorname{MSE}_{\mathcal E}},
\end{equation}
where $D$ is the number of scalar RGB values per frame. Prefix LPIPS averages
framewise LPIPS over $[g(j),b)$ against the clamped reference. It measures
reference agreement and does not establish exact preservation of the
original decoded prefix.

\paragraph{Boundary error and aggregation.}
Boundary error compares generated and reference frame changes at the first
editable decoded frame,
\begin{equation}
E_{\mathrm{boundary}}
=\frac{1}{D}\left\|(x_b-x_{b-1})
-(x_b^\star-x_{b-1}^\star)\right\|_1.
\end{equation}
The preceding frame belongs to committed history for $\mathrm{CST}_{R}$
and to the prefix for $\mathrm{CST}_{T}$. This metric measures agreement with
the reference transition, rather than smoothness alone. We compute one score
per group and method, then report the median across groups for LPIPS, PSNR,
and boundary error. Camera-trajectory and HY-WorldPlay benchmark results use
means. The overall median is computed from individual groups, not from
subgroup medians.

\paragraph{Camera trajectories.}
We align estimated and requested camera-to-world poses at common frame
indices and express each trajectory relative to its first pose. Let
$(\widehat R_i,\widehat t_i)$ and $(R_i,t_i)$ denote these aligned poses.
We normalize estimated translation by
\begin{equation}
 k=\arg\max_i\|t_i\|_2,\qquad
 \alpha=\frac{\|t_k\|_2}{\|\widehat t_k\|_2}.
\end{equation}
Alignment and scale estimation use the full aligned trajectory before
selecting the scored pose indices $\mathcal P$. Following
\citet{hunyuanworld2025hy}, $R_{\mathrm{dist}}$ averages geodesic rotation
error in radians and $T_{\mathrm{dist}}$ averages squared Euclidean
translation error between $\alpha\widehat t_i$ and $t_i$. We then average
these per-example scores. The scored interval contains four poses for
$\mathrm{CST}_{R}$ and $4-m$ suffix poses for $\mathrm{CST}_{T}$, unlike
the published 61-frame evaluations.

\paragraph{Timing and computation.}
The $K-r$ remaining solver evaluations exclude corrector execution,
state reconstruction, context-cache processing, and decoding. Cache rebuilding
requires additional transformer work, so this count does not measure total
transformer calls or runtime. Reference construction is excluded from CST timing.

\subsection{Ablation Details}
\label{app:ablation_details}

The ablations use separately trained HY-WM1.5 correctors and the same
matched evaluation protocol as the full model. Unmatched supervision assigns
each source capture a target from a different prompt through a fixed,
seeded permutation within its data partition. Source inputs are retained,
and the target is taken at the same solver step. This breaks the
counterfactual pairing without mixing training and validation prompts.

The mask ablation disables both the mask input channel and multiplication
of the predicted residual by the temporal mask during training and
inference. The suffix mask still selects the elements used in the matching
loss. Prefix coordinates can therefore change without direct prefix
supervision. This ablation tests the combined effect of mask conditioning
and output restriction. It does not isolate output masking alone.

Action error measures disagreement between estimated and requested relative
camera motions. After the alignment above, let $\Delta\widehat R_i$ and
$\Delta R_i$ be successive relative rotations, and $\Delta\widehat t_i$ and
$\Delta t_i$ the corresponding translations in the preceding camera frame.
Let $\theta(Q)$ denote the geodesic angle of rotation $Q$ in radians.
For increments entering the scored pose interval $\mathcal P$, we compute
\begin{equation}
 E_{\mathrm{action}}=\frac{1}{|\mathcal P|}\sum_{i\in\mathcal P}
 \left[
 \frac{\theta(\Delta\widehat R_i\Delta R_i^\top)}{\pi/60}
 +\frac{\|\alpha\Delta\widehat t_i-\Delta t_i\|_2}{0.08}
 \right].
\end{equation}
The normalizers are the command magnitudes of $3$ degrees and $0.08$ scene
units. We include the increment from the preceding pose into the first
scored pose, giving four increments for $\mathrm{CST}_{R}$ and $4-m$ for
$\mathrm{CST}_{T}$. Table~\ref{tab:actionsplice_ablations} reports the median
per-example action error. Its fidelity and boundary metrics use the
references and aggregation defined above.

\section{Additional Results and Human Evaluation}
\label{app:additional_experiments}
\subsection{Receipt-Step Sensitivity}
\label{app:interrupt_sweeps}

\begin{table}[!htbp]
    \centering
    \caption{Receipt-step sweep for $\mathrm{CST}_{R}$ on HY-WM1.5
    with $K=4$. Metrics follow Appendix~\ref{app:evaluation_protocol}.}
    \label{tab:hy_receipt_sweep}
    \small
    \resizebox{\linewidth}{!}{%
    \begin{tabular}{cccccccc}
        \toprule
        Receipt step $r$ &
        \makecell{Remaining\\solver evaluations} &
        \makecell{Pixel-ready\\latency (ms) $\downarrow$} &
        LPIPS $\downarrow$ &
        PSNR $\uparrow$ &
        \makecell{Boundary\\error $\downarrow$} &
        $R_{\mathrm{dist}}\downarrow$ &
        $T_{\mathrm{dist}}\downarrow$ \\
        \midrule
        1 & 3 & 7578.8 & 0.0643 & 23.13 & 0.0204 & 0.108 & 0.024 \\
        2 & 2 & 7124.0 & 0.0831 & 20.51 & 0.0196 & 0.048 & 0.025 \\
        3 & 1 & 6026.6 & 0.2694 & 18.39 & 0.0246 & 0.180 & 0.064 \\
        \bottomrule
    \end{tabular}%
    }
\end{table}
Later receipt reduces latency, but $r=3$ has higher LPIPS and boundary error
and lower PSNR than $r=1$ or $r=2$.
\subsection{Temporal-Boundary Sensitivity}
\label{app:boundary_sweep}

\begin{table}[!htbp]
    \centering
    \caption{Temporal-boundary sweep for $\mathrm{CST}_{T}$ on
    HY-WM1.5 at $r=2$. Scored intervals, references, and metrics follow
    Appendix~\ref{app:evaluation_protocol}.}
    \label{tab:hy_boundary_sweep}
    \small
    \resizebox{\linewidth}{!}{%
    \begin{tabular}{ccccccc}
        \toprule
        Boundary $m$ &
        \makecell{Prefix\\LPIPS $\downarrow$} &
        \makecell{Suffix\\LPIPS $\downarrow$} &
        \makecell{Suffix\\PSNR $\uparrow$} &
        \makecell{Boundary\\error $\downarrow$} &
        $R_{\mathrm{dist}}\downarrow$ &
        $T_{\mathrm{dist}}\downarrow$ \\
        \midrule
        1 & 0.0403 & 0.0701 & 24.74 & 0.0189 & 0.062 & 0.047 \\
        2 & 0.0443 & 0.0597 & 25.53 & 0.0208 & 0.049 & 0.093 \\
        3 & 0.0395 & 0.0578 & 24.69 & 0.0230 & 0.049 & 0.023 \\
        \bottomrule
    \end{tabular}%
    }
\end{table}
The scored suffix shortens as $m$ increases, so values across boundaries
refer to different temporal regions.

\subsection{Directed Action-Transition Breakdown}
\label{app:transition_breakdown}

\begin{table}[!htbp]
    \centering
    \caption{Action-transition breakdown for $\mathrm{CST}_{R}$ on
    HY-WM1.5 at $r=2$. Each row reports mean camera errors and median LPIPS
    under Appendix~\ref{app:evaluation_protocol}. The final row pools all
    30 examples.}
    \label{tab:hy_transition_breakdown}
    \small
    \setlength{\tabcolsep}{10pt}
    \begin{tabular}{lccc}
        \toprule
        Action transition &
        $R_{\mathrm{dist}}\downarrow$ &
        $T_{\mathrm{dist}}\downarrow$ &
        \makecell{LPIPS to\\rollback $\downarrow$} \\
        \midrule
        Forward $\rightarrow$ Backward & 0.002 & 0.028 & 0.0479 \\
        Forward $\rightarrow$ Yaw-left & 0.078 & 0.031 & 0.1285 \\
        Backward $\rightarrow$ Forward & 0.003 & 0.043 & 0.0593 \\
        Backward $\rightarrow$ Yaw-left & 0.054 & 0.003 & 0.1042 \\
        Yaw-left $\rightarrow$ Forward & 0.074 & 0.028 & 0.1434 \\
        Yaw-left $\rightarrow$ Backward & 0.079 & 0.019 & 0.2654 \\
        \midrule
        All transitions & 0.048 & 0.025 & 0.0831 \\
        \bottomrule
    \end{tabular}
\end{table}

Performance varies by transition, with the highest rollback-relative LPIPS
for yaw-left to backward.

\subsection{Human Annotation Interface}

Figure~\ref{fig:annotation_interface} shows the blinded annotation interface.
Randomized sample identifiers hide method identity. For each action segment,
annotators record whether the requested action is visible in the editable
region and count decoded frames from the start of that region to the first
visible response. The scored region is the active chunk for
$\mathrm{CST}_{R}$ and its suffix for $\mathrm{CST}_{T}$. We report action
following as a percentage and stale frames as a mean.

\noindent\begin{minipage}{\linewidth}
    \centering
    \includegraphics[width=0.9\linewidth,trim=0 220 0 0,clip]{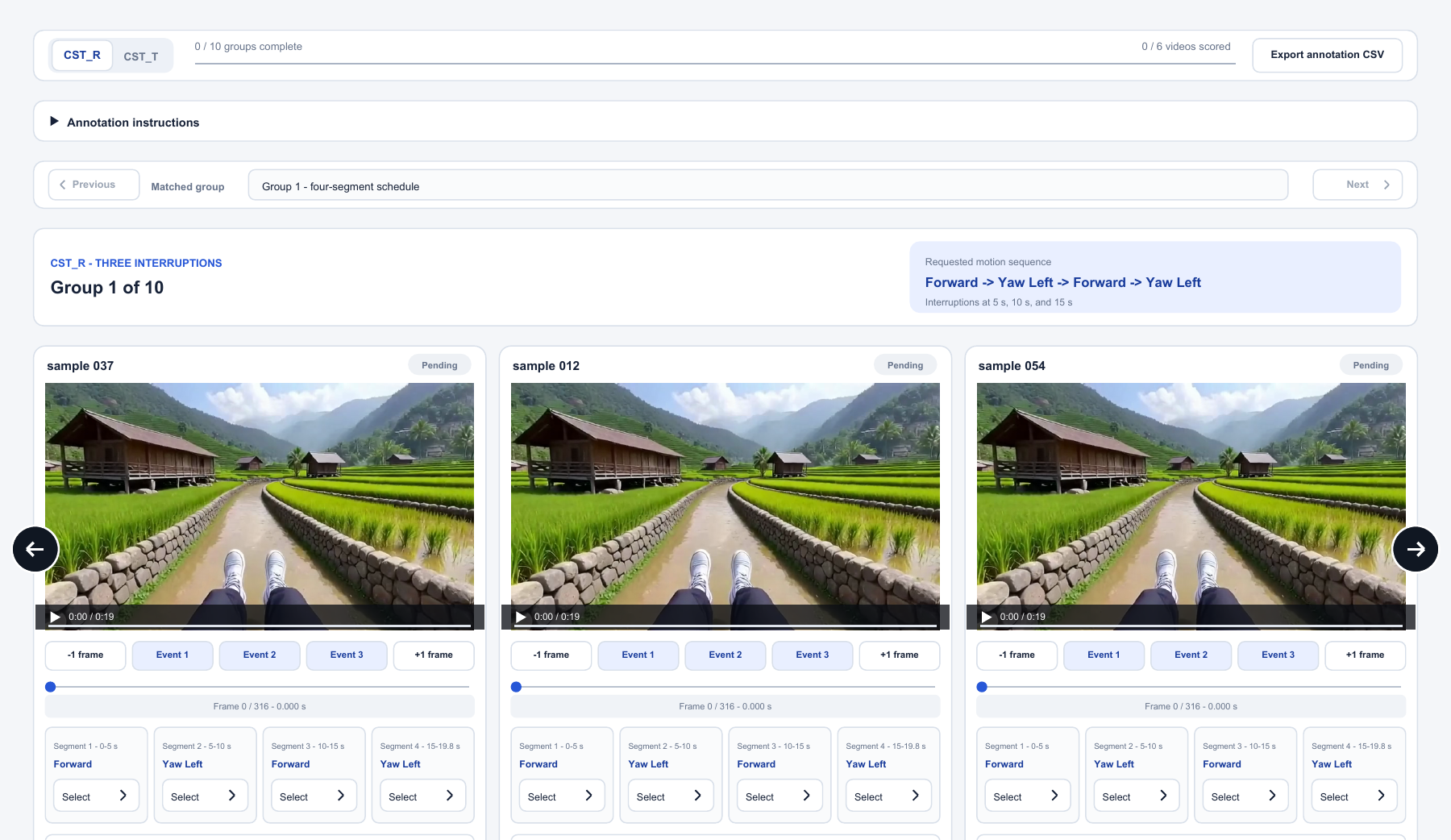}
    \captionof{figure}{Blinded action-response annotation interface. Randomized
    sample identifiers conceal method identity, and segment controls support
    frame-level inspection around each action update.}
    \label{fig:annotation_interface}
\end{minipage}

\end{document}